\documentclass[10pt,twocolumn,letterpaper]{article}

\usepackage[dvipsnames]{xcolor}
\usepackage{iccv}
\usepackage{times}
\usepackage{epsfig}
\usepackage{graphicx}
\usepackage{amsmath}
\usepackage{amssymb}
\usepackage{booktabs}
\usepackage{pgfplots}
\usepackage{tikz}
\usepackage{adjustbox}
\newcommand{\YSR}[1]{{\color{red} {\bf} YSR: #1}}

\usepackage[pagebackref=true,breaklinks=true,letterpaper=true,colorlinks,bookmarks=false]{hyperref}

\iccvfinalcopy 

\usepackage[capitalize]{cleveref}
\crefname{section}{Sec.}{Secs.}
\Crefname{section}{Section}{Sections}
\Crefname{table}{Table}{Tables}
\crefname{table}{Tab.}{Tabs.}

\def\iccvPaperID{7077} 

\ificcvfinal\fi

\begin{document}

\title{Efficiently Robustify Pre-Trained Models}

\author{Nishant Jain \thanks{Correspondence to Nishant Jain at njain@cs.iitr.ac.in.}\\
IIT Roorkee\\
\and
Harkirat Behl\\
Microsoft Research\\
\and
Yogesh Singh Rawat\\
CRCV, UCF\\
\and
Vibhav Vineet\\
Microsoft Research\\
}

\maketitle
\ificcvfinal\thispagestyle{empty}\fi

\begin{abstract}
      A recent trend in deep learning algorithms has been towards training large scale models, having high parameter count and trained on big dataset. However, robustness of such large scale models towards real-world settings is still a less-explored topic.
      In this work, we first benchmark the performance of these models under different perturbations and datasets thereby representing real-world shifts, and highlight their degrading performance under these shifts. We then discuss on how complete model fine-tuning based existing robustification schemes might not be a scalable option given very large scale networks and can also lead them to forget some of the desired characterstics. Finally, we propose a simple and cost-effective method to solve this problem, inspired by knowledge transfer literature. It involves robustifying smaller models, at a lower computation cost, and then use them as teachers to tune a fraction of these large scale networks, reducing the overall computational overhead.
      We evaluate our proposed method under various vision perturbations including ImageNet-C,R,S,A datasets and also for transfer learning, zero-shot evaluation setups on different datasets. Benchmark results show that our method is able to induce robustness to these large scale models efficiently, requiring significantly lower time and also preserves the transfer learning, zero-shot properties of the original model which none of the existing methods are able to achieve.

\end{abstract}

\section{Introduction}
\label{sec:intro}

\begin{figure}
    \centering
    \includegraphics[width=\linewidth]{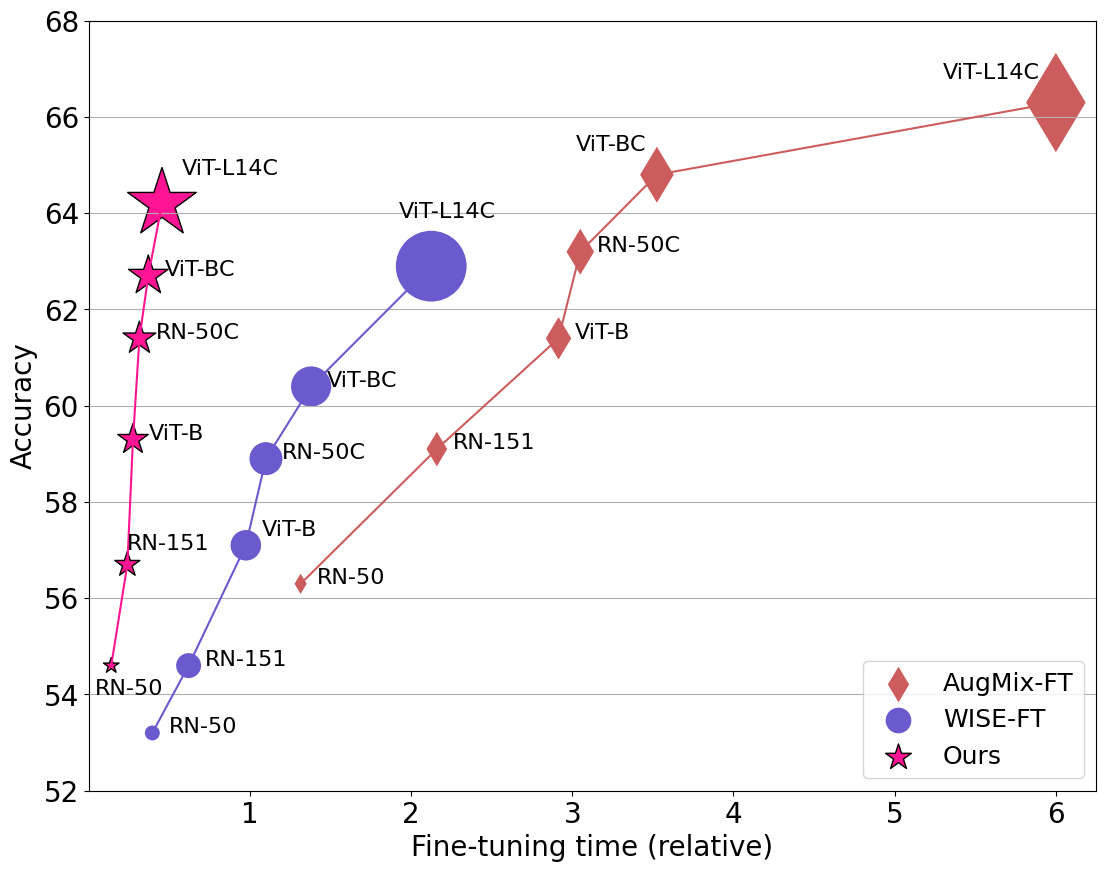}
    \caption{ImageNet-C accuracy v/s training time comparison.
    Our method is on the pareto-front (achieves better robust accuracy in much lesser time) compared to the state-of-the-art methods Augmix based Complete fine-tuning and WISE-complete fine-tuning. The data points labelled with suffix "C" correspond to CLIP models.}
    \label{fig:pareto_front}
\end{figure}

Large scale deep neural networks trained on large scale data have revolutionized the modern AI era. They are significantly effective in solving practical problems of high importance.
%
%
These include object detection, zero-shot classification, image segmentation, image generation, and many other applications \cite{khan2018review,pathak2018application, radford2021learning, zhai2022lit,dosovitskiy2010image, tran2022plex, yang2022unified,he2016deep}.
%

Though the large models have shown impressive results on many vision problems \cite{radford2021learning, tran2022plex}, their reliability under distribution shift e.g., under illumination changes, geographical variations, camera properties etc., is still under-explored.
%
%
In this paper, we fist investigate the behavior of large  models under distribution shifts. We analyse popular models under synthetic perturbations to images \cite{hendrycks2019benchmarking}, natural distribution shifts \cite{hendrycks2021many, hendrycks2021natural}, differently styled images \cite{wang2019learning} and dataset shift \cite{barbu2019objectnet}. %
Our analysis of models of various sizes, architecture families (transformers or CNNs) and training modalities (uni or multi-modal) establishes their brittleness under distribution shifts.
%
%

This analysis begs the question: can we induce robustness to large vision models without sacrificing their original properties?
It is critical to simultaneously maintain \textit{clean} accuracy on the original datasets, improve \textit{robust} accuracy on the shifted data and preserve the \textit{transfer learning} capabilities of the large models. Further, computation efficiency during both training and inference is beneficial.
%
%
%
%
%
%

While several prior works can be used to make large-scale models robust, they do not possess the desired properties discussed above. One direction involves fine-tuning the model \cite{wortsman2022robust, jia2022visual}. This generally suffers from either poor performance under synthetic perturbations or requires significant training time. 
%
Another line of work could be to use advanced augmentation techniques (e.g., aug-mix, pix-mix) \cite{hendrycks2022pixmix, hendrycks2019augmix, hendrycks2021many, devries2017improved}. They are effective under synthetic perturbations and natural shifts in the data. However, they require significantly larger compute time and lead to the large models forgetting their original and transfer learning properties. Figure \ref{fig:pareto_front} shows this analysis in a pareto-front plot for two of the recently proposed robustness methods.
%
%
%


To this end, we propose 
a knowledge transfer method to induce robustness to large models that possesses all the desired properties discussed above. 
It makes large models robust efficiently (refer Fig.\ref{fig:pareto_front}). 
%
%
%
We take a \textit{plug-and-play} approach: insert an additional small robust module and update only a very small portion of the existing large models.
%
%
%
%
To achieve robustness, we explore a new direction: a relatively much smaller but robust model inducing robust knowledge to a large model. Though this provides a novel look at the knowledge distillation approach, a straight-forward application leads to the large models forgetting their original properties. 
For this challenging task of ensuring that clean accuracy is preserved in the clean module, robustness induced into the robust module and correct module selected at test time, we propose a novel uncertainty-aware knowledge distillation technique. This allows us to fulfil all our required objectives.
Since our method involves updating only a small chunk of the large network, it achieves low training latency (refer section \ref{sec:exp}). To the best of our knowledge, this is the first time such a setup has been used involving knowledge transfer from a smaller to a large model. Further, it should be noted that smaller models can be made robust by using prior works like advance augmentation methods \cite{hendrycks2019augmix, hendrycks2021many, hendrycks2022pixmix}.

We evaluate our method under various distribution shift on ImageNet data \cite{ILSVRC15} in section \ref{sec:exp}. It includes ImageNet-C \cite{hendrycks2019benchmarking}, ImageNet-R \cite{hendrycks2021many}, ImageNet-A \cite{hendrycks2021natural}, ImageNet-sketch \cite{wang2019learning}, ImageNet-V2. We also evaluate on ObjectNet \cite{barbu2019objectnet} and its perturbed variations ObjectNet-C. 
We show results for both multi-modal (various CLIP models) and unimodal (various architectures including both ResNets and Vision Transformers). Alongside this, we also test our method on other datasets in the transfer learning setup, to analyze further if the desired properties of the model are restored.
%
In all these cases, our method outperforms prior approaches on robust accuracy while still performing at par on clean accuracy. At the same time, possessing desired characteristics like transfer learning capabilities (refer section \ref{sec:exp}) and being efficient during training and inference.
\section{Related Work}
\paragraph{Large scale models.}

In recent years, studies \cite{zhai2022scaling, yuan2021tokens,he2022masked} have shown that training large models such as vision transformers \cite{dosovitskiy2010image} on large datasets can improve accuracies significantly.
Several works \cite{paul2022vision,bhojanapalli2021understanding} have evaluated these models for robustness lately . 
Furthermore, these large models can been trained either in a unimodality setup 
\cite{paul2022vision} or multi modality setup
\cite{radford2021learning, zhai2022lit, yang2022unified}. %
Though they achieve good performance on several downstream tasks, any modification of these large models can lead to forgetting of the knowledge contained in them. In contrast, we propose a method that allows to adapt the model parameters without sacrificing their properties.
\vspace{-3mm}
\paragraph{Finetuning and Augmentation based methods to achieve robustness.} Several advanced augmentation techniques have been proposed to improve robustness of the deep learning based models. Examples includes cut-out, cutmix, mixup, augmix, pixmix and others \cite{devries2017improved, yun2019cutmix, hendrycks2019augmix, zhang2017mixup}. 
%
%
Further, a recently proposed method WISE  interpolates the fine-tuned and original model parameters \cite{wortsman2022robust} for robustness \cite{zhang2017mixup}. 
%
Generally these techniques are a computation heavy process and also leads to modifying the main network parameters that could lead to these large models forgetting in their original properties.
In our approach, we use some of these advanced augmentation technique to make our teacher network robust. We ensure that our robust approach does not sacrifice the large models' properties and is also computationally efficient.

\vspace{-4mm}
\paragraph{Knowledge distillation}
It involves transferring knowledge from a large network to a smaller network by minimizing the distance between the predicted logit distribution of the student and teacher networks \cite{hinton2015distilling}.
It proved to be highly effective on standard downstream datasets \cite{alkhulaifi2021knowledge}, 
 \cite{gou2021knowledge}. 
%
In all these KD applications, knowledge is transferred from a larger network to a smaller network. In contrast, we propose a method to induce robustness to a larger network by transferring knowledge from a smaller (teacher) network.

\vspace{-2mm}
\section{Robustness Analysis}
\label{sec:benchmark}
\vspace{-1.5mm}

In this section, we analyze the image classification performance of models of different shapes and sizes, with different training settings (unimodal or multimodal). 
We stress test the models under both synthetic and natural perturbations.
Especially, contrasting the behaviour of multimodal models (vision-language) \textit{vs.} unimodal (image only).

\begin{figure*}
    \centering
    \includegraphics[width=0.33\linewidth]{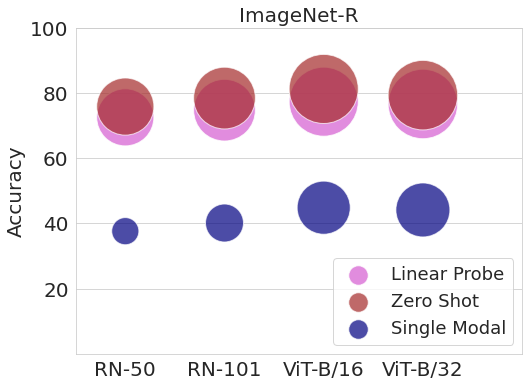}
    \includegraphics[width=0.33\linewidth]{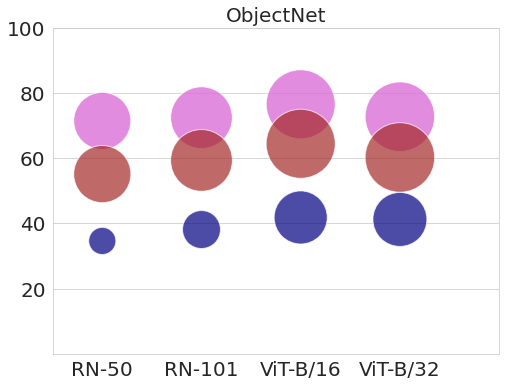}
    \includegraphics[width=0.33\linewidth]{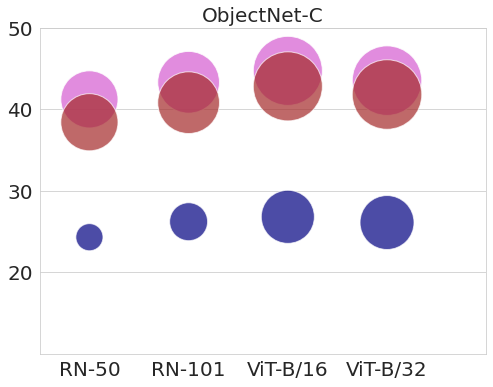}
    \caption{Analysis of multi-modal linear-probe, multi-modal zero-shot and unimodal networks under various distribution shifts including ImageNet-R, ObjectNet, ObjectNet-C. The x-axis denote the model architecture and y-axis denotes the accuracy.
    }
    \vspace{-2mm}
    \label{fig:bench_lp}
\end{figure*}

\textit{Models.} In unimodal setting, we analyse Resnet-50, ResNet-101 and ResNet-150 \cite{he2016deep}, ViT-small, ViT-base and ViT-large models \cite{dosovitskiy2010image} trained on ImageNet \cite{ILSVRC15}.
In multimodal setting, we analyse CLIP \cite{radford2021learning} model with backbones including ResNets: CLIP-RN50 and CLIP-RN101, and transformers: CLiP-ViT B/16, CLiP-ViT B/32. We also analyze self-supervised unimodalities trained on large datasets as against only ImageNet pretrained ones. For this, we use the masked autoencoders \cite{he2022masked} and DINO V2 \cite{oquab2023dinov2} models proposed recently, shown to be highly effective in representation learning. We analyze two architecture, ViT-B/16, ViT-B/32 for MAE and ViT-B/14 for DINO V2.

\noindent
\textit{Datasets.} 
We evaluate the models on various shifted version of ImageNet \cite{ILSVRC15}: ImageNet-Corrupted (ImageNet-C) \cite{hendrycks2019benchmarking} , ImageNet-Rendition (ImageNet-R) \cite{hendrycks2021many} and ImageNet-Sketch (ImageNet-S) \cite{wang2019learning} datasets.
For natural shifts, we use ImageNet-Adversarial (ImageNet-A) comprising natural adversarial examples\cite{hendrycks2021natural} (results in supplementary). 
Further, we also evaluate the models on ObjectNet \cite{barbu2019objectnet} and its perturbed version ObjectNet-C, which we generate by applying all the 15 corruptions of ImageNet-C to ObjectNet data. 

\begin{figure*}
    \centering
    \includegraphics[width=0.31\linewidth]{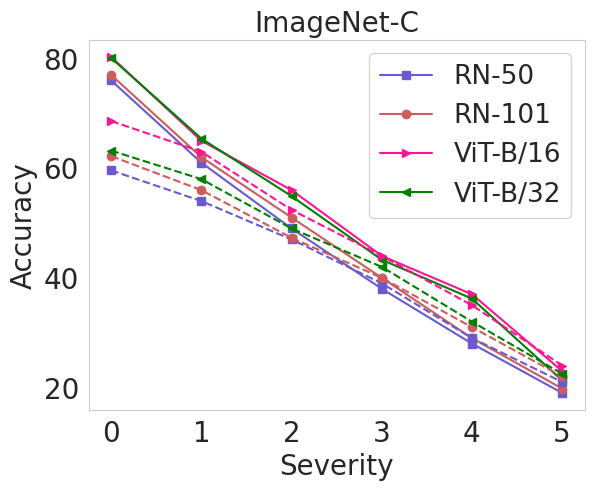}
    \includegraphics[width=0.31\linewidth]{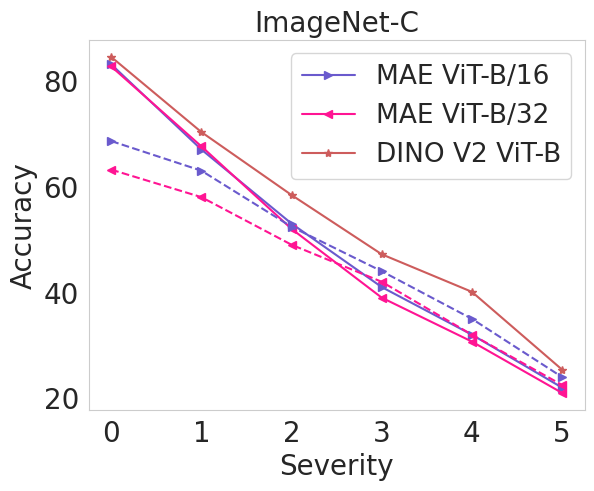}
    \includegraphics[width=0.32\linewidth]{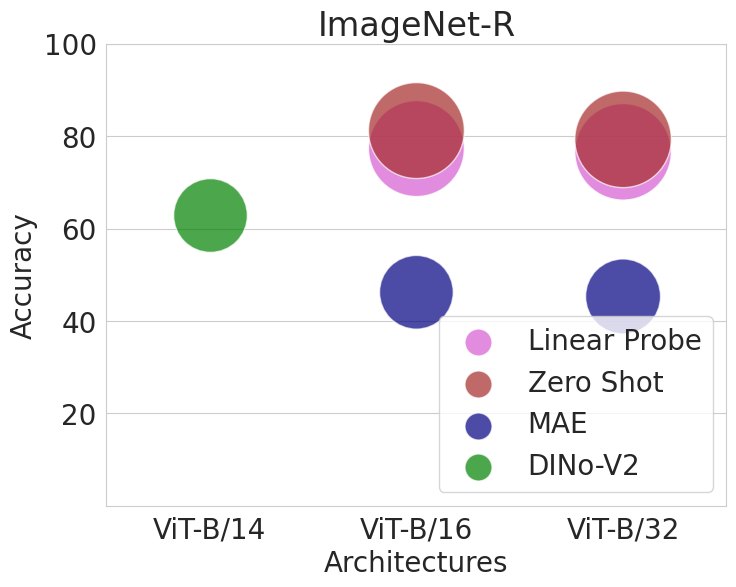}
    \caption{\textbf{Left: }Comparison of accuracy score (y-axis) on ImageNet-C dataset against various severity levels (x-axis) of perturbations, including both Unimodal (solid line) and Multi-Modal CLIP (dashed line) architectures.  Unimodal architectures are ImageNet-pretrained and Multi Modal architectures correspond to Zero-shot CLIP. \textbf{Mid:} Comparison of Self-Supervised unimodalities and CLIP Zero-Shot models on the ImageNet-C benchmark. \textbf{Right:} Comparison of self-supervised unimodalities and CLIP Linear Probe and Zero Shot models on the ImageNet-R dataset.
    }
    \vspace{-2mm}
    \label{fig:im_c_benchmark}
\end{figure*}

\noindent
\textit{Experimental Setup.} In the unimodal case, we evaluate ImageNet trained models from timm \cite{rw2019timm} on ImageNet-C, ImageNet-R, ImageNet-S (supplementary), ObjectNet and ObjectNet-C datasets. 
Further, we evaluate the multimodal models in the linear-probe \cite{radford2021learning} and zero-shot settings. For the linear probe setup, please refer to the CLIP paper. It is done on the ImageNet dataset and then evaluated on all of these.

%

\noindent
\textit{Results.} 
Fig. \ref{fig:im_c_benchmark} (left) shows analysis of all the architectures on ImageNet-C data with varying severity levels, for ImageNet pretrained unimodalities (solid lines) and Zero-Shot CLIP-models (dashed lines). 
They all suffer similarly when severity is increased and even though start from different accuracies, converge to similar values at high severity. This implies under high perturbations they break-down equally. However, the robustness of CLIP based models is slightly higher. One possible reason is that they start from lower values of clean accuracy attributed to their zero-shot nature. 

\noindent
Fig. \ref{fig:bench_lp} shows the analysis of various CLIP model architectures on the ImageNet-R, ObjectNet and ObjectNet-C datasets under both Linear Probe and zero-shot settings along with the unimodal (ImageNet pretrained) counterparts. 
For the linear probe setting, the models maintain accuracy on ImageNet-R and ObjectNet datasets whereas suffer significantly on ObjectNet-C. On the other hand, zero shot models show better accuracy on the ImageNet-R (compared to ImageNet), slightly lower on the ObjectNet dataset and suffer on the ObjectNet-C dataset similar to linear probe setting. From the results, it can be observed that zero-shot CLIP is more robust on ObjectNet-C than linear probe CLIP based on the relative drop on accuracy. Also, the zero-shot CLIP model outperforms the linear probe one on the ImageNet-R dataset, even though the linear probe was on ImageNet.
\noindent
 For unimodal case, all models observe significant drop in accuracy and perform poorly (much worse than CLIP Linear Probe and Zero-Shot) under all the shifts.
 

\noindent
{\textit{Self Supervised Unimodalities}.  We further analyse another case where the unimodal models are trained in a self supervised fashion on large datasets as against the ImageNet pretrained models. For this, we use the masked autoencoders \cite{he2022masked} and DINO V2 \cite{oquab2023dinov2} models proposed recently, shown to be highly effective in representation learning. Fig. \ref{fig:im_c_benchmark} (mid and right) provides their robustness analysis on the ImageNet-C,R datasets alongside the CLIP models. For ImageNet-C, similar to Fig. \ref{fig:im_c_benchmark} (left), these models also converge similar to the multi-modal models at the highest severity levels and their relative drop from severity level 1 to 5 is higher. On ImageNet-R, again, multi-modal models perform significantly better than the uni-modal models. These observations are similar to the case seen in Fig.\ref{fig:bench_lp}.} 


\noindent
\textit{Empirical Conclusion.} From all the plots, it can be observed that multi-modal networks are much better than the unimodal counterpart on ImageNet-R, ObjectNet and ObjectNet-C datasets and can be seens as more robust for these settings. However, they also see a significant drop in accuracies under the perturbations in ImageNet-C, especially at higher accuracy levels. Again all of the architectures used show similar drops in accuracy. Also, zero-shot multi-modal networks seem to be more robust than their linear probe counterpart in the presence of distribution shifts (comparing the drop in accuracy from ImageNet to other datasets). On the architectural side, transformer models appear to be more robust than the ResNets for both single and multi-modalities, given their higher accuracy.

\begin{figure*}
    \centering
\includegraphics[width =0.9\linewidth]{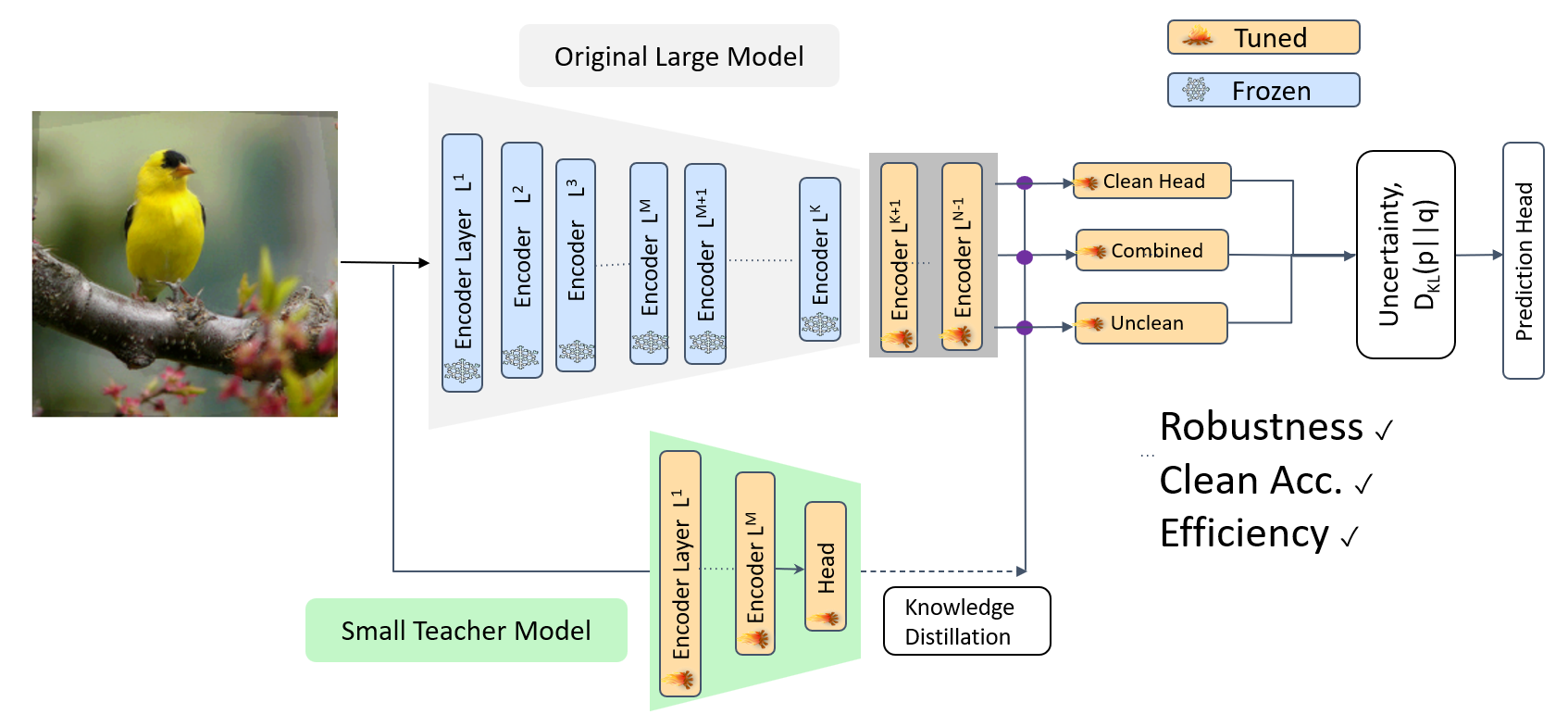}
    \caption{The end-to-end workflow of our proposed method. It involves firstly robustifying a small teacher model using advanced augmentation methods (lower stream). Then using this model along with augmented data, it tunes a small chunk of the large-scale (student) model. As described in sec. \ref{sec:method}, we add two more heads to the student model resulting in total three heads. Yellow colored encoder denotes the (only few) tunable layers and blue colored encoders correspond to frozen layers.
    Finally, at the inference time, the head used for prediction is selected via estimating uncertainty in predictions and analyzing KL divergence between the
    distributions predicted by each head. For more details, please refer section \ref{sec:method}.
    }
    \label{fig:pipeline}
\end{figure*}

\section{Methodology}
\label{sec:method}
\noindent \textbf{Problem Description.} The goal of our work is to make large pre-trained computer vision models robust without sacrificing their original properties. 
This is critical because we want models to work well both in in-domain and out-of-domain settings. 
Computational efficiency is also important because pre-training already requires massive amounts of compute, so efficient techniques are more versatile.

\noindent
A popular technique to robustify a model is to use advanced augmentation techniques like aug-mix \cite{hendrycks2019augmix} or pix-mix \cite{hendrycks2022pixmix}. 
This involves fine-tuning model parameters using the augmented dataset and is very effective at improving robustness.
However, such fine-tuning is sub-optimal, as models could forget their original representation properties and at the same time require large computation resources. \\
\noindent \textbf{Method Overview:}
To this end, we propose a novel Plug-and-Play method. 
First, alongside the original clean classification head, we plug a robust and a combined head into the model. 
Second, we induce robustness from a \textit{small robust teacher} ({here the small is relative to the large pretrained model}) into the robust head.
However, this leaves the challenging task of ensuring that clean accuracy is preserved in the clean head and robustness induced into the robust head. More importantly, we need to ensure that these heads can be correctly selected at test time.
Third, we propose a novel uncertainty-aware knowledge distillation technique, which allows us to fulfil all our required objectives.
The proposed method is discussed below and also shown in the Fig. \ref{fig:pipeline}.
\subsection{Augmented Architecture}
\noindent
Let us denote the original model as $\mathcal{M}$. The network can be seen as made of three components:
$\mathcal{M}_b$ the \textit{backbone} network spanning from initial layer to the $K^{th}$ layer, $\mathcal{M}_s$ the \textit{shared tunable section} spanning $(K+1)^{th}$ layer to $(N-1)^{th}$ layer, and $\mathcal{M}_h$ the \textit{prediction head section} from $N^{th}$ layer till the end (refer figure \ref{fig:pipeline}). 
Thus the overall network can written as:
\begin{equation}
\mathcal{M}=\mathcal{M}_h\circ\mathcal{M}_s\circ\mathcal{M}_b ,
\end{equation}
where $\theta$, $\theta_b$, $\theta_s$ and $\theta_h$ denote the respective component parameters. \\
To address the issue of robustness transfer with preservation of desired characteristics like clean accuracy, transfer learning, etc., we plug-in two more prediction head sections on top of the shared tunable section. This results in a total of three classification heads as shown in figure \ref{fig:pipeline}. 

\subsection{Robustness Distillation from a Small Teacher}
\noindent
At the core of our approach lies use of a knowledge distillation (KD) framework to induce robustness to a large network. 
In standard KD, knowledge is usually transferred from a large network to a small network. 
\textit{Au contraire}, we provide a novel view of KD. We show that robustness can be transferred from a small robust model to large models.
\noindent
For the small robust teacher (denoted as $\mathcal{M}_t$), we take small image-classification models and robustify them using standard techniques (a combination of augmentation techniques AugMix \cite{hendrycks2019augmix} and DeepAugment \cite{hendrycks2021many}). 
A \textit{small} teacher is essential for an efficient method.

\subsection{Uncertainty Aware Knowledge Distillation}
\noindent
While we have introduced a robust head and plan to induce robustness from the small model. It is a challenging task to ensure that clean head preserves clean accuracy, robust head learns robustness and the heads are appropriately selected at test time. We next discuss a novel strategy to achieve these goals.

\noindent
We update the parameters of shared-tunable $\theta_s$ and prediction sections $\theta_h$, keeping the backbone network frozen as shown in figure \ref{fig:pipeline}. 
We use the same augmented training data here as used for robustifying the small (teacher) model. It is denoted as $\mathcal{D}^a$ and contains both clean $(x^{c}, y)$ and augmented samples $(x^{a},y)$ samples. 
Given the augmented training data and the robust network $\mathcal{M}_t$, the parameter estimation for the model can be written as:
\vspace{-2mm}
\begin{equation}
\vspace{-2mm}
    \{\theta_s,\theta_h\} \sim \mathcal{P}(\{\theta_s,\theta_h\}|\theta, \mathcal{M}_t, \mathcal{D}^{a}).
\end{equation} 
\subsubsection{Generalized distillation}
\vspace{-1.5mm}
We next discuss our strategy to optimize for both knowledge (clean accuracy) and robustness distillation (robust accuracy).
Note that the teacher for the clean head is a copy of the initial large model to preserve the original properties. 
This estimation is done using a weighted combination of classification loss $\mathcal{L}_c(x,y,\theta)$ and distillation loss \cite{hinton2015distilling} $\mathcal{L}_d(\theta_T,\theta_S,x)$, where $(x,y)\in D^a$, $\theta$ denotes parameters of the prediction network, $\theta_T$ denotes teacher model parameters, $\theta_S$ denotes student model parameters. 

The head section (with parameters $\theta^c_h$) corresponding to the \textit{clean} head is updated only using the clean examples.
The \textit{combined} head ($\theta_m$) uses both clean and unclean examples and the \textit{unclean} head ($\theta_u$) uses only the augmented examples. 
Thus, for clean examples in a randomly sampled batch of data, the set of updated parameters due to clean head section prediction is 
$\theta_c^c=\{\theta_s,\theta_h^c\}$  and combined head section is $\theta_m^c=\{\theta_s,\theta_h^m\}$.  
Similarly for augmented examples, updated parameter set due to unclean head section prediction is $\theta_u^u=\{\theta_s, \theta_h^u\}$ and combined head section is $\theta_l^u=\{\theta_s,\theta_h^m\}$. 

\noindent
Thus, the final loss function to update w.r.t. clean examples is (denoted as $\mathcal{L}_{clean}$):
\vspace{-1mm}
\begin{equation}
\vspace{-2mm}
\begin{aligned}
     \mathcal{L}_{c}(x,y,\theta_c^c) + \mathcal{L}_{d}(x,\theta_c^c,\theta_t) + \mathcal{L}_{c}(x,y,\theta_m^c) + \mathcal{L}_{d}(x,\theta_m^c,\theta_t),
\end{aligned}
\end{equation}
and similarly for unclean examples (denoted as $\mathcal{L}_{aug}$):  
\vspace{-2mm}
\begin{equation}
\begin{aligned}
     \mathcal{L}_{c}(x,y,\theta_u^u) + \mathcal{L}_{d}(x,\theta_u^u,\theta_t) + \mathcal{L}_{c}(x,y,\theta_m^u) + \mathcal{L}_{d}(x,\theta_m^u,\theta_t).
\end{aligned}
\end{equation}
Finally, the cost function $\mathcal{L}$ for a given batch of data can be written as:
\begin{equation}
    \mathcal{L} = \beta\mathcal{L}_{clean} + (1-\beta)\mathcal{L}_{aug},
\end{equation}
where $\beta=1$ for clean and $\beta=0$ for unclean examples. 
%
%

\vspace{-2mm}
\subsubsection{Uncertainity aware head selection} 
\vspace{-1.5mm}
We need a reliable head selection, such that clean head is selected for clean examples (to preserve clean accuracy) and unclean head for shifted examples (robustness).
\vspace{-1.5mm}
\paragraph{Uncertainty.} Modelling uncertainty in predictions corresponding to each of the heads can be a way to select the final head as the most certain head. Clean head should be the most certain on clean examples and similarly unclean for augmented examples. For this, we use Dropout \cite{gal2016dropout} as it gives allows \textit{Bayesian} approximation.
At training time, we update the tunable portion of $\mathcal{M}$, starting from encoder $L^{K+1}$ in Fig.\ref{fig:pipeline} using \textit{dropout} regularization. This can be done by just setting the dropout rate to some non-zero fraction in the existing implementation. This dropout is a part of all the three heads. 

\noindent
{At the inference time, we activate the dropout, for each of the heads, to estimate a predictive distribution from the model as against a point estimate \cite{gal2016dropout,kendall2017uncertainties}. We take $K$ forward passes through each of these heads and then for each head we calculate mean and std of the outputs and use them as the mean and std of the predicted distribution from the model. This is referred as Monte Carlo Dropout. Finally, we use std directly as the measure of uncertainty ($\mathcal{U}_{mc}$).}


\paragraph{KL Divergence.} Now, there can be a case where the clean model completely breaks down for a noisy input and predicts a random output with very low $\mathcal{U}_{mc}$ (highly certain). Given the test-distribution is unknown, this can be a case with significant probability. To handle this, we also calculate the distance between the predicted distributions of each of the clean and unclean head with the combined head using KL divergence only at Inference time. 
This results in the following objective for selecting the final prediction head $h_f$:
\begin{equation}
    h_{f} = \arg \min_{{k}\in \{c,u\}} \mathcal{U}_{mc}\cdot\text{KL}(\phi^m_{l}(x)||\phi^k_{l}(x)),
    \label{eq:head_select}
\end{equation}
where $h_c$, $h_m$, $h_u$ correspond to clean, combined, unclean heads respectively and  
$\phi^c_{l}$, $\phi^m_{l}$, $\phi^u_{l}$ are the corresponding prediction functions. 
{Thus, the desired head out of clean/unclean is selected using eq. \ref{eq:head_select}. Note, the third head is here just to select the correct desired head for the input from the clean and unclean head via the KL divergence term in the head selection metric in eq. \ref{eq:head_select}.}
{In supplementary, we have provided a detailed ablation on the utility of each of these components and also the comparison against a naive confidence based head selection baseline.}
\subsection{Zero-Shot Multi-Modal scenario} 
\noindent
The method described above can be directly applicable to uni-modalities and vision encoders of multi-modalities by attaching a classification head, similar to the Linear Probe setup \cite{radford2021learning}. We further adapt our scheme to zero-shot setup for these multi-modal networks which comprise both vision and text encoders. For this, we apply our scheme in the presence of text encoder and use the dot products between the vision encoder embeddings ($\phi_v(x)$) and the prompt embedding obtained from the text encoder ($\phi_{text}(Y)$, where $Y$ denotes the set of prompts corresponding all classes present in the data) , $\phi_l(x) = \phi_v(x)\cdot\phi_{text}(Y)$, as the model prediction for both classification and distillation losses.




\begin{table*}[t]
    \centering
    \begin{adjustbox}{width=\textwidth}
    \begin{tabular}{lc|cccccc|ccccc|c|c}
    \toprule
       & & \multicolumn{6}{c|}{{Distribution Shifts}} & \multicolumn{5}{c|}{{Transfer Learning}}&  Avg. &{Latency} \\
       & IN & IN-V2 & IN-R & IN-Sketch & ObjectNet & IN-A  & IN-C & Tiny-IN & Flowers & Places & iNaT & SUN  &shifts &(ms/img)\\  
    \midrule
    {CLIP ViT-L/14@336px} & & & & & & & & &\\
    Zero-Shot \cite{radford2021learning}  & 76.2  & 70.1  & 88.9  & 60.2  & 70.0 $\pm$ 0.2 &  77.2  &60.6  & \textbf{85.2 }  & \textbf{98.8}  & \textbf{74.87}  & \textbf{68.20}  & \textbf{82.20}  &71.7  & -\\
    Fine-Tuning (LP) \cite{radford2021learning} & 85.4 & 75.8 & 84.1 & 57.4 & 66.3 & 75.3 & 57.9 & \textbf{85.2} & \textbf{98.8} & - & -& -&69.5&320\\
    Fine-Tuning & 86.1 & 76.6 & 79.7 & 57.7 & 63.4 & 65.4 & 52.1 & {83.5} & 97.6 & 72.12 & 65.70 & 78.89 &65.8 &2500\\
    VPT \cite{jia2022visual} & 85.8 & 74.2 & 80.1 & 56.9 & 63.9 & 66.1 &52.6& {85.2} & 98.6 & - & -& -&65.5&500\\
    WISE-FT (LC) \cite{wortsman2022robust} & 85.1 & 76.6 & 85.2 & 63.0 & 71.0 & 79.5 &62.1 & {85.1} & 98.4 & 73.98 & 67.45 & 81.12 &72.9&570\\
    WISE-FT(E2E) \cite{wortsman2022robust} & \textbf{86.9} & \textbf{79.5} & \textbf{90.1} & 65.0 & 72.1  & {80.6} & 62.9 & 83.4 & 97.4 &72.94 & 66.28 & 80.23&75.0&3950\\
    \midrule
    K.D. from ViT-B/16 teacher  & & & & & & & & & \\
    Single-Teacher & 84.7 & 76.2 & 88.2 & 61.1 & 68.7& 76.3&  60.9& 83.3& 97.1 & - & -& -& 71.9&530\\
    Only K.D. & 85.1 & 77.3 & 89.3 & 63.9 & 70.6 & 78.6& 61.7& {85.1}&\textbf{98.8} & - & - & -& 73.5&620\\
    Ours  & 85.4 & 79.1& 89.9& \textbf{65.8}& \textbf{73.2}& \textbf{{80.9}}& \textbf{64.9}& \textbf{85.2}& {98.7} & 74.69 & \textbf{68.20} & 82.10 & \textbf{75.6}&750\\
     \bottomrule
     
    \end{tabular}
    \end{adjustbox}
    \vspace{0.1in}
    \caption{ \textbf{Visual Evaluations results.}
    Comparison with existing robustification methods and complete fine-tuning on various distribution shift and transfer learning benchmarks for CLIP model. The column \textit{Avg. shifts} shows the average accuracy overall the six distribution shifts. {Last column shows training cost per image for all the models. The numbers reported are average over five runs.}
    }
    \label{tab:main_table}
\end{table*}

\vspace{-1mm}
\section{Experiments}
\label{sec:exp}



\noindent
We evaluate the presented approach in making large models robust on the benchmark datasets and perturbations described in the section \ref{sec:benchmark} for image classification task. We show performance on clean accuracy, robust accuracy and transfer learning properties on downstream  datasets.

%
\vspace{-1.5mm}
\paragraph{Experimental Setup.} We demonstrate results of our approach under two settings. First corresponds to using visual encoders of uni or multi-modal models (as described in linear probing approach in the Sec. \ref{exp_mm}) and attaching a classification head to the visual encoders. We term this as the  Visual Evaluation or {\em VE} setup.
%
Next, we also provide results (Sec. \ref{exp_mm}) in multi-modal models settings where both text and vision are used in the zero-shot (or {\em ZS}) setting under dataset shift. 
Along with clean and robust accuracy, we also compare different methods on whether they can preserve the transfer learning capabilities of the large models. 
The robust teacher for both settings is a single modal network trained by finetuning complete model using augmentation based techniques using a combination of augmix and deep augmentation techniques (as described in Sec. \ref{sec:method}).
%
%

%
%
%
%
%
\vspace{-2mm}
\paragraph{Datasets.} We evaluate the presented approach on the ImageNet validation set and its perturbed variations, ImageNet-C 
 \cite{hendrycks2019benchmarking}, ImageNet-R \cite{hendrycks2021many}, ImageNet-S \cite{wang2019learning} and ImageNet-A \cite{hendrycks2021natural} for robust analysis in the VE setup. Further, we use the ObjectNet data \cite{barbu2019objectnet} and its  perturbed variation ObjectNet-C (Corrupted) version for the zero-shot evaluation tasks or ZS setup.
For transfer learning experiment, we show results {five datasets (Tiny-ImageNet \cite{le2015tiny}, Flowers \cite{Nilsback08}, PLACES025 \cite{li2017learning}, iNaturalist2021 \cite{van2021benchmarking} and SUN397 \cite{xiao2010sun})} in the VE setup. {For the ZS setup, we instead show results on dataset shift on} six datasets (Cars \cite{krause20133d}, Flowers \cite{Nilsback08}, CIFAR100 \cite{krizhevsky2009learning}, SUN397 \cite{xiao2010sun}, ObjectNet \cite{barbu2019objectnet}), {where the zero-shot model is directly evaluated on these datasets}. More information about these datasets have been provided in the supplementary material.
\vspace{-4.5mm}
\paragraph{Baselines and metrics.} We now describe baselines used for the VE and ZS setups. For the VE experiments involving only the visual encoders, we compare against five prior approaches. First approach involves adapting the the same number of parameters as in the presented linear probe approach (Sec. \ref{sec:benchmark}). Second baseline involves naive finetuning full network using the current dataset. Third baseline is the visual prompt tuning approach \cite{jia2022visual} that involves adapting input prompt parameters while fixing the feature network parameters. 
We also compare against the recently proposed WISE \cite{wortsman2022robust} framework for finetuning large networks. 
Finally, we define a new baseline, Augmentation based Partial Fine-Tuning or {\em APT}
to show the effectiveness of multi-headed scheme. It involves directly updating the small part of the large network (same number of tunable layers as ours), using the augmentation based technique. 

%

We further define two more baselines, as variations of our proposed scheme, to further highlight its importance. The first baseline, \textit{Only K.D.}, involves doing knowledge distillation directly from the Small Teacher Network (STN) teacher to Large Learner Network (LLN) student without using the proposed multi-headed architecture and copy of initial LLN as teacher for clean examples. The second baseline, \textit{combined head}, involves using copy of initial LLN as teacher for clean examples and STN for augmented, but without the multi-headed architecture, \textit{i.e.} requiring only the combined head.

For the ZS setting, the set of baselines involves the existing zero-shot multi-modal network, along with the APT baseline (ZS-APT) and complete fine-tuning baseline (ZS-Full tuning). Both these baselines are tuned similar to our method, in presence of the text encoder, as described in section \ref{sec:method} (Zero Shot Multi-Modal scenario). 

%
We use the accuracy metric on clean and perturbed dataset to evaluate each of the methods. Furthermore, we also calculate the robustness metrics for evaluating under perturbations, proposed for the ImageNet-C dataset in the supplementary. 

\begin{table*}[]
    \centering
    \begin{adjustbox}{width=\textwidth}
    \begin{tabular}{lc|cccccc|ccccc|c}
    \toprule
       & & \multicolumn{6}{c|}{{Distrbution Shifts}}&  \multicolumn{5}{c|}{{Dataset Shift}} & Avg. \\
       & IN & IN-V2 & IN-R & IN-S & ON & IN-A  & IN-C  & Cars & Flowers & CIFAR100 & SUN397 & ON-C&shifts\\  
    \midrule
    {CLIP ViT-L/14@336px} & & & & & & & & & & \\
    Zero-Shot  & 76.2 & 70.1 & 88.9 & 60.2 & 70.0 & 77.2 &60.6& {78.8} & \textbf{{78.3}} & 77.5 &{68.4} & 52.2&71.1\\
    ZS-Full Tuning & \textbf{86.5} & \textbf{77.1} & 88.2 & 58.9 & 65.5 & 78.2 & 53.3 & {{76.3}} & {76.8} & 77.1 & 67.2&  51.5&70.0\\
    \midrule
    ZS-APT  & 84.8 & 76.3 & 89.2 & 60.7 & 68.8 & 77.9 & 62.3 & {77.2} & 77.9 & 77.3 & 67.9& 54.3&71.8\\
    Ours (ViT-B/16 teacher) & {86.3} & 76.8& \textbf{90.2}& \textbf{62.9}& \textbf{{71.6}}& \textbf{74.6}&\textbf{78.9}& \textbf{78.9}&{78.1} & \textbf{77.6} & \textbf{69.3} & \textbf{58.2}&\textbf{73.6}\\
     \bottomrule
     
    \end{tabular}
    \end{adjustbox}
    \vspace{5pt}
    \caption{ \textbf{Zero Shot results.} Comparison with existing robustification methods and complete fine-tuning on various distribution and dataset shifts under the zero shot setup using CLIP model. Last column shows the average accuracy including all the dataset and distribution shifts. {All these results are a result of zero-shot evaluation of the robustly tuned classifier model using only the ImageNet-1K dataset. No further tuning on any of the other datasets. The numbers reported are average over five runs.}}
    \label{tab:main_table_mm_zs}
\end{table*}

\begin{table*}[t!]
    \centering
    \begin{adjustbox}{width=\textwidth}
    \begin{tabular}{lc|cccccc|cc|c|c}
    \toprule
       & & \multicolumn{6}{c|}{{Distrbution Shifts}} & \multicolumn{2}{c|}{{Transfer Learning}} & Avg. &{Latency} \\
       & IN & IN-V2 & IN-R & IN-Sketch & ObjectNet & IN-A  & IN-C & Tiny-IN & Flowers & shifts &(ms/img)\\  
    \midrule
    {ViT-L/14} & & & & & & & & &\\
    Standard Training & 82.8 & 75.3 & 49.4 & 40.4 & 45.2 & 51.9 & 51.5 & {{84.5}} & {98.1} & 52.3&2200\\
    VPT \cite{jia2022visual} & 82.1 & 74.4 & 47.2 & 38.1 & 45.3 & 51.2 & 50.3 & {84.1} & 97.9 & 51&420\\
    WISE-FT (LC, optimal $\alpha$) \cite{wortsman2022robust}& 82.6 & 76.2 & 54.2 & 48.3 & 49.2 & 55.3 &54.2 & {84.0} & 98.1 & 56.2&500\\
    WISE-FT (E2E, optimal $\alpha$) \cite{wortsman2022robust}& \textbf{83.6} & \textbf{78.4} & {58.3} & 52.1 & \textbf{52.3}  & {57.7} & 55.1 & 83.8 & 97.5& 58.9&3400\\
    \midrule
    K.D. from ViT-Small teacher & & & & & & & & & \\
    Single-Teacher & 81.9 & 75.9 & 52.7 & 46.3 & 47.4& 53.2&  53.3& 83.8& 97.2&54.8 & 470\\
    Only K.D. & 82.2 & 76.7 & 53.4 & 46.9 & 48.1 & 54.2& 53.2& {84.3}&{97.9} & 55.4&550\\
Ours & 82.7 & 78.2& \textbf{59.1}& \textbf{52.7}& {51.6}& \textbf{58.5}& \textbf{58.3}& \textbf{84.5}& \textbf{98.1} &\textbf{59.7}& 700\\
     \bottomrule
     \vspace{0.1in}
    \end{tabular}
    \end{adjustbox}
    \vspace{-2.5mm}
    \caption{\textbf{Unimodal results.} Comparison with existing robustification methods and complete fine-tuning on various distribution shift and transfer learning benchmarks for single-modal ViT-L/14 model, pretrained on JFT dataset. 
    Second last column shows average accuracy over the distribution shifts and the last one shows training latency per image. The numbers reported are average over five runs.}
    \vspace{-2mm}
    \label{tab:main_table_single_modal}
\end{table*}

\subsection{Evaluating Multi Modality Trained Methods} 
\label{exp_mm}
\paragraph{Visual Evaluation.} Table \ref{tab:main_table} shows the comparison of our method with the baselines under various distribution shifts for ImageNet dataset along with the transfer learning capabilities and training time per image. 
%
It uses the CLIP ViT-L/14@333px model for all the methods. Our method uses ViT-B/16 as the teacher model.
 It can be observed that 
even though WISE-E2E performs best for two shift scenarios, it suffers from high training time and poor performance under transfer learning tasks (accuracy drop of 1.8 and 1.4) which is a bottleneck with E2E fine-tuning methods. On the other hand, the methods like VPT which require low training time perform poorly under distribution shift (average accuracy drop greater than 5\% when compared with Zero-Shot model). On the other hand, performs best on four  distribution shift tasks  boosting up accuracy upto 2\% (ImageNet-C) and is able to achieve the same accuracy as the zero-shot model on the transfer learning experiments and also in a significantly lesser time compared to WISE-E2E (approximately 5 times faster).

\begin{figure*}
    \centering
    \includegraphics[width=\linewidth]{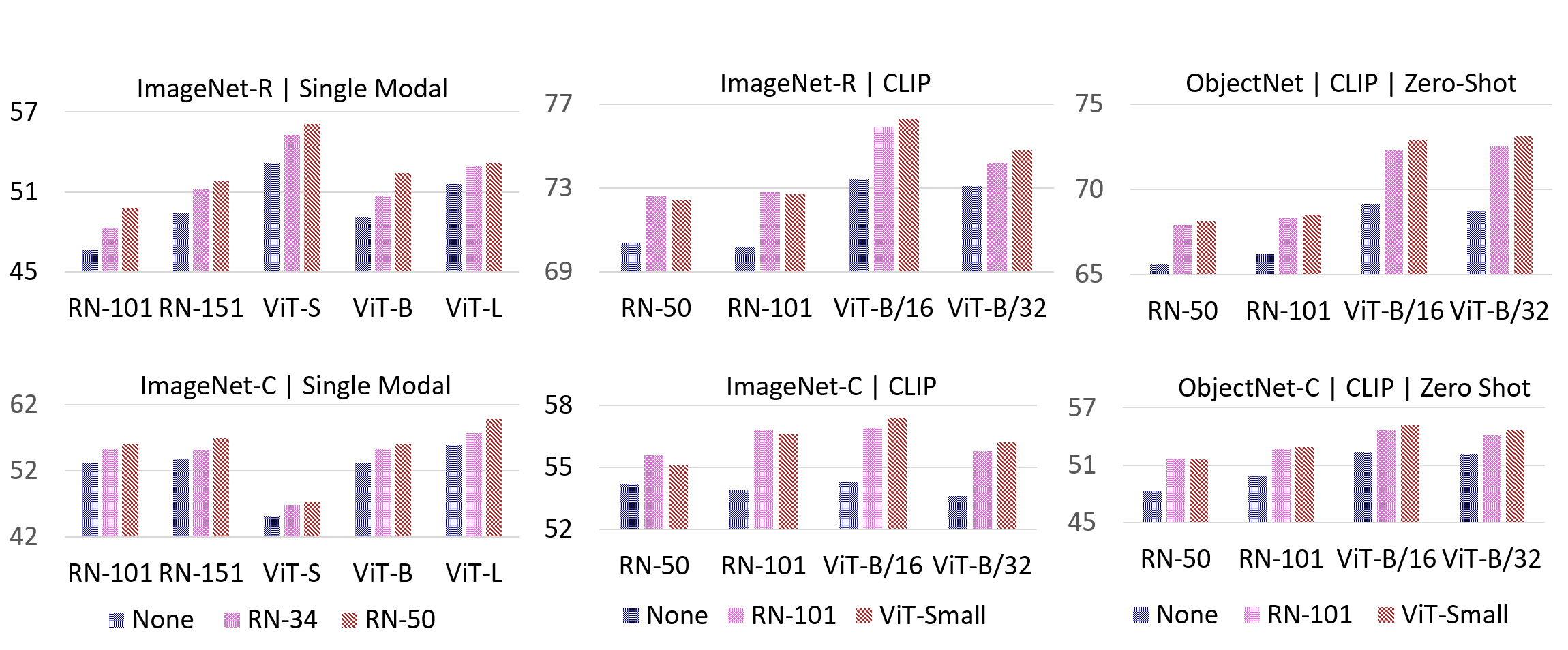}
    \caption{Analysis of our methods for various Teacher (RN-34, RN-50)-Student pairs and the APT baseline (None) applied to students. For Zero-Shot, None corresponds to the given zero-shot model without any tuning. x-axis: Students, y-axis: Accuracy, vertical bars: Teacher. }
    \label{fig:result_hist}
\end{figure*}

We further evaluate our method against the APT baseline, we defined, for various model architectures as Teachers and Learners. 
Figure \ref{fig:result_hist} (top left and bottom left) shows this analysis, with visual encoder of four CLIP models (RN-50, RN101, ViT-B/16 and ViT-B/32) as students and single modal networks as teacher evaluated on ImageNet-R and ImageNet-C datasets. The rows with teacher as \textit{None} correspond to the APT baseline. 
For ImageNet-C, accuracy is improved by more than 3\% for most cases, and is atleast 2.5 \% for all cases. This knowledge transfer doesn't rely much on the architectural family of student or teacher as only marginal difference is there in the improvements offered by ViT or ResNet architecturs as teachers on CLIP ViT/ResNet students (less than 0.3\% difference observed when updating CLIP RN-50 with ResNet-101 or ViT-Small). 


%
For ImageNet-R, our method provides gains 2.0\% for most cases with maximum as 2.9\%, compared to APT baseline.
For accuracy of the teacher models, please refer supplementary.

\vspace{-5mm}
\paragraph{Zero Shot.} We now apply our scheme on multi-modal CLIP networks using both text and visual encoder for a complete zero-shot setup, as discussed in section \ref{sec:method}. Table \ref{tab:main_table_mm_zs} shows the results for our method and baselines under this setup, for various distribution shifts to the ImageNet data and also for zero-shot evaluation on new datasets (dataset shift). Here again, ViT-B/16 is used as the teacher for our method. It can be observed that our method shows best performance under all the distribution shifts with minimum difference of 1\% (IN-R) and maximum of 4.5\% (IN-C). Also, it is the best performing zero-shot method for four out of five dataset shifts with the maximum improvement being on ObjectNet-C 3.9\%). This shows that it improves zero-shot properties of the model as compared to the complete fine-tuning which instead degrades the performance under dataset shift as observed in table \ref{tab:main_table_mm_zs}.
Table \ref{fig:result_hist} (top mid and bottom mid) shows the further accuracy analysis under this setting for various student (single modal) and teacher (CLIP) pairs on ImageNet-R, ImageNet-C datasets. 
%
The evaluation is done on the ObjectNet dataset and its perturbed version  ObjectNet-C.  
Here, the baseline (rows with teacher as \textit{None}) is  just the zero-shot large-scale network (using text and visual encoder) and doesn't use the tuning set. Again, our scheme improves accuracy by a significant amount, atleast 2.6 percent under zero-shot setting for the dataset shift. 
Again it works irrespective of the architecture or modality of the teacher network (eg. CLIP RN-50 student and RN-101, ViT-S teachers in the figure). \\
We further analyze the effect of our method and the APT baseline on the transfer learning setup on tiny-ImageNet and Flowers dataset for various teacher-student pairs in the supplementary.

\noindent
{\textbf{Inference time Latency}. Since our method attaches extra heads, although quite small, on top of the existing model, we also analysed the inference time overhead it adds on top of the naive model. We observed that this overhead was quite small. For instance, CLIP ViT-L/14 model used in table \ref{tab:main_table}, GFLOPs for baselines is 81.17 and ours is 83.7, (3\% overhead). }

\vspace{-2mm}
\subsection{Evaluating Unimodally Trained Methods}
\vspace{-1mm}
\noindent
We now analyze our method for a unimodal \textit{Learner} network scenario, comparing it with all the baselines on different distribution shifts/transfer learning tasks, as done for the multi-modal VE setup.
Table \ref{tab:main_table_single_modal} shows this comparative  results. Here again, our method emerges as the winner for four out of the six distribution shifts with minimum gain of 0.8\% and maximum gain of 3.2\%. Furthermore, it is also able to preserve transfer learning capabilities, as shown in table \ref{tab:main_table_single_modal} for {Tiny-IN, Flowers, PLACES205, iNaturalist2021, SUN397 datasets},  of the original model whereas other baselines (VPT and WISE) suffer.

\noindent
Figure \ref{fig:result_hist} (last column) shows this comparison for various Leaner-Teacher pairs consisting both ViT and ResNet architectures.
Again, the rows with Teacher \textit{None} correspond to the APT baseline. 
%
For majority cases on ImageNet-C, our method improves accuracy by greater than 3 percent, when compared to the baseline.
Similarly, on the ImageNet-R dataset, it shows greater than 5\% for most cases with maximum going to 3.2\% for RN-50 teacher and RN-101 student.
 Furthermore, increasing the size of teacher model (from RN-34 to 50) results in improved performance. 
 Finally, both our method and the baseline we compare to, make significant improvement in performance on the perturbed datasets (especially ImageNet-C), compared to the ImageNet pretrained models (refer section \ref{sec:benchmark} for performance of these ImageNet-pretrained models).

\vspace{-1.5mm}
\subsection{Ablations}
\vspace{-1mm}
\paragraph{Robustification schemes.} We now compare the effect of using a different robustification schemes for the teacher model, used in our method. We limit ourselves to augmentation based robustification schemes such as AugMix, PixMix, DeepAugment \textit{etc.}. 
Table 
\ref{tab:ab_robust_sm} shows the results for this comparison for the single modal setup when a ResNet-101 student model is using ResNet-50 teacher. 
Using Pix-Mix or combining with AugMix does improves the accuracy by around 1-1.5 over our current technique (Augmix+DeepAugment). 
Our scheme shows that
that gains in teacher model can be efficently transferred to the student. 
\vspace{-5mm}
\paragraph{Amount of parameters tuned.} We further analyze the effect of tuning different proportions of the LLN using our scheme. It can be observed that tuning more parameters increases the accuracy on ImageNet-C and R datasets at the cost of clean accuracy on ImageNet dataset. 

\noindent
Please refer to the supp. material for more description about the ablations for understanding model design choices (KL Div., Uncertainty, Multiple-heads) and other details.


\begin{table}
    \centering
    \scriptsize
    \begin{tabular}{c|c|cc}
        \toprule
        Method & ImageNet & ImageNet-C & ImageNet-R \\
        \midrule 
         Augmix \cite{hendrycks2019augmix} & 78.3 & 55.2 & 49.1 \\
         Augmix+DA \cite{hendrycks2021many} & 78.9 & 56.1 & 49.8 \\
         PixMix \cite{hendrycks2022pixmix} & 79.1 & 56.8 & 49.9 \\
         AugMix+PixMix & 79.0 & 57.1 & 50.9 \\
         \bottomrule
    \end{tabular}
    \vspace{0.5mm}
    \caption{\small Ablation of various robustification schemes used to robustify the teacher used in our approach. Here, a ResNet-101 is the student and ResNet-50 is the teacher.}
    \label{tab:ab_robust_sm}
\end{table}
\begin{table}
    \centering
    \scriptsize
    \begin{tabular}{c|c|cc}
        \toprule
        Fraction-tuned & ImageNet & ImageNet-C & ImageNet-R \\
        \midrule 
          0   & 77.5 & 39.2 & 40.1 \\
         0.05  & 77.4 & 52.9 & 46.5 \\
         0.10  & 77.2 & 55.1 & 47.7 \\
         0.20  & 76.4 & 57.2 & 48.8 \\
         0.25 & 76.2 & 57.4 & 49.2 \\
         0.30 & 75.9 & 57.5 & 50.1 \\
         \bottomrule
    \end{tabular}
    \vspace{0.5mm}
    \caption{\small Ablation showing variation in model performance as a function of its proportion of tuned parameters using our method, Here, ResNet-101 is the student and ResNet-50 is the teacher.}
    \label{tab:params_tuned}
\end{table}

\section{Conclusion}
\noindent
We first benchmark and discuss existing large pre-trained models under various shifts to the data. Following this, we proposed an  effective method to distill robustness to these networks via small robust models, at the same time preserving the characterstics of the large models. 
Results on various distribution shift settings showed that our method is effective and efficient in making large pretrained models robust to several distribution shifts, and also retaining their transfer learning properties.
\\
\noindent
\textbf{Limitations.} Though we have provided extensive empirical evidence to demonstrate the benefit of our approach, a theoretical underpinning is missing. We leave theoretical analysis as an interesting future work.

{\small
\bibliographystyle{ieee_fullname}
\bibliography{egbib}
}

\clearpage

\appendix

\section{Further Ablations}
\label{supp_ablation}
We further provide an analysis of various variants, so as to understand importance of each of the proposed modules namely the \textit{multi-headed architecture}, \textit{knowledge distillation} from small to big network and \textit{uncertainty/kl divergence} used at the inference time. We begin with analyzing the importance of proposed inference procedure.
\subsection{Inference}
We first begin with ablation on the proposed scheme for inference involving  \textit{Monte-Carlo Dropout} (MCD) uncertainty $\mathcal{U}_{mc}$ and KL divergence calculation (Sec. \textcolor{red}{4.3.2} in the paper). We also define a new term along with accuracy, for analyzing these inference time ablations. It is the fraction of examples in the test set, for a given dataset , which are assigned the correct head (clean for in-distribution and unclean for distribution/dataset shift). We denote it by $F_{correct}$.
We start by analyzing the importance of KL divergence and thereby comparing our proposed scheme against the variant without any KL divergence calculation at the inference time. 
\\
\textbf{KL Divergence}
Table \ref{tab:ablation_table} shows the analysis of our method with and without (\textit{w/o}) KL divergence calculation at the inference time, on all the distribution shift datasets used in the paper for Visual Evaluation setting. The variant without the KL divergence term at the inference time is denoted as \textit{Ours w/o KLD}  in the table.
It also shows the results for transfer learning experiments.
The student model corresponds to multi-modal CLIP \textit{ViT-L@333px} network and a single modal ViT-B/16 is used as the teacher.  The last column of the table consists of average accuracy across all the distribution shift datasets.
It can be observed that there is a small gain  (average of 0.9\%) across all the distribution shift scenarios when using our complete method as compared to this variant. Thus, using KL divergence at the inference time further rectifies the model performance. Also, Table \ref{tab:ablation_table_fraction} shows the analysis of $F_{correct}$ metric on all the distribution shift datasets used. Here also, a noticeable difference is observed implying KL divergence is significantly helping in deciding the correct head for the input. Next, we consider the ablation for the $\mathcal{U}_{mc}$ term.
\\
\textbf{Uncertainty}. Similar analysis for $\mathcal{U}_{mc}$ is shown in Table \ref{tab:ablation_table} and the row corresponding to \textit{Ours w/o $\mathcal{U}_{mc}$} denotes this case. Again, removing the uncertainty component causes a depreciation in performance, but more than removing the KL divergence component (average decrease of 1.9\% as compared to 0.9\% from removing KL Divergence term at the inference time). Hence, both components are necessary for most optimal prediction. Furthermore, from Table \ref{tab:ablation_table_fraction} showing analysis of $F_{correct}$, it can be observed that using this $\mathcal{U}_{mc}$ term is significantly helping in deciding the correct head, again impacting more than the KL Divergence term. \\
{\textbf{Confidence based head selection.} We also compare our head selection scheme with a case where instead of KL divergence or uncertainty, the predictive confidence of the clean and unclean heads is used for selecting the final classification head. Table X also contains the results for this method in the row \textit{Ours(max logit)}. Again our proposed head selection scheme surpasses it with a significant margin.
}

\subsection{Knowledge Distillation}
We now discuss the importance of the knowledge distillation (KD) module (Sec. \textcolor{red}{4.3} in the paper) proposed in our work, used for tuning the student model parameters. For this, we define another variant of our method with the multi-headed architectural scheme but without any KD. Table \ref{tab:ablation_table} also shows the analysis for this case using same models and scenarios as the above cases, compared against our method. The row corresponding to \textit{Ours w/o K.D.} corresponds to this ablation.
It can be observed that our method improves performance significantly as compared to this ablation (average 2.5\% across distribution shifts), thereby showing utility of knowledge transfer. 
On clean dataset, only a marginal increase in performance is observed. Thus, the knowledge distillation component in our method plays a significant role in inducing robustness.
Further comparison against this ablation for different teacher student pairs is shown in table \ref{tab:ab_kd}.
Again significant performance improvement, upto 2.6\%, can be observed for our method using knowledge distillation.

\subsection{Amount of dataset}
\begin{figure}
    \centering
    \includegraphics[width=0.9\linewidth]{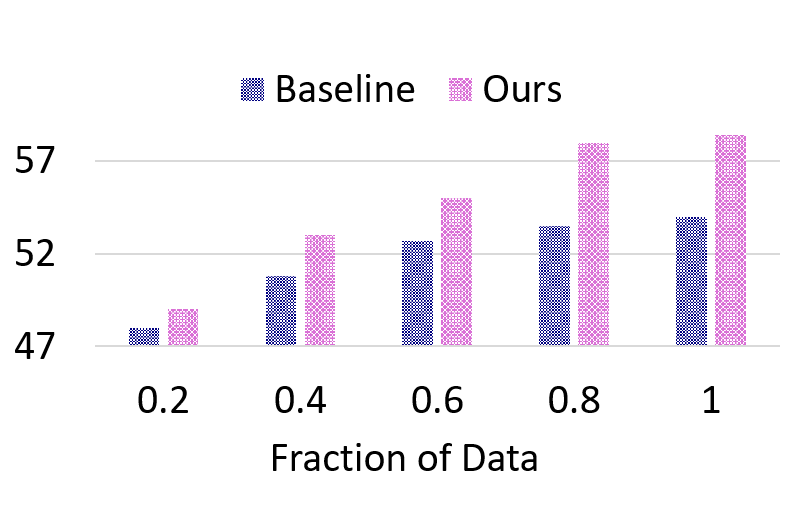}
    \caption{Ablation on amount of distillation data v/s accuracy. The x-axis shows the fraction of augmented data used in distillation and the y-axis shows the accuracy achieved for each fraction by the APT baseline and our method on the ImageNet-C dataset. Here, ResNet-101 is used as the student network, updated using a ResNet-34 teacher, both being single modal.}
    \label{fig:amount_of_data}
\end{figure}

We further analyze the impact of using different amounts of data while distilling knowledge, as it is crucial in deciding the computational overhead. 
Let us denote the total number of examples in the augmented data as $N^{'}$ and the number of examples (sampled randomly) for updating the student model as $aN^{'}(0<a<1)$.
Here, we analyze our scheme for different values of $a$ when it is applied on the ResNet-101 student with ResNet-34 as the teacher. The results are shown in Figure \ref{fig:amount_of_data}. It can be observed that as the data for tuning increases, the performance gap between our method and baseline  increases, implying robustness of our scheme increases significantly w.r.t. data as compared to baseline.

\begin{table*}[t]
    \centering
    \footnotesize
    \begin{tabular}{lc|cccccc|cc|c}
    \toprule
       & & \multicolumn{6}{c|}{{Distribution Shifts}} & \multicolumn{2}{c|}{{Transfer Learning}}&  Avg.  \\
       & IN & IN-V2 & IN-R & IN-Sketch & ObjectNet & IN-A  & IN-C & Tiny-IN & Flowers  &shifts \\  
    \midrule
    \multicolumn{1}{l}{CLIP \textit{ViT-L/14@336px}}\\
    \midrule
    \multicolumn{2}{c}{}&\multicolumn{6}{c}{{Inference Time Ablations}}\\
    \midrule
    \textit{Ours w/o KLD} & 84.7 & 78.3 & 89.3 & 65.6 & 72.8& 79.3&  63.1& 85.2& 98.7 & 74.7\\
    \textit{Ours w/o $\mathcal{U}_{mc}$} & 83.9 & 77.7 & 88.4 & 64.9 & 71.5 & 77.6& 62.2& {85.2}&{98.7} & 73.7\\
    \textit{Ours (max logit)} & 84.2 & 77.6 & 88.1 & 64.1 & 71.5 & 79.2& 62.9& {85.2}&{98.7} & 73.7\\
    \midrule
    \multicolumn{2}{c}{}&\multicolumn{6}{c}{{Knowledge Distillation Ablation}}\\
    \midrule
    \textit{Ours w/o K.D.} & {84.9} & {77.6} & {86.8} & 63.2 & 70.3  & {78.3} & 62.1 & \textbf{85.4} & \textbf{98.8} &73.1\\
    \midrule
    \textit{Ours}  & \textbf{85.4} & \textbf{79.1}& \textbf{89.9}& \textbf{65.8}& \textbf{73.2}& \textbf{{80.9}}& \textbf{64.9}& {85.2}& {98.7} & \textbf{75.6}\\
     \bottomrule
     
    \end{tabular}
    \vspace{0.1in}
    \caption{ \textbf{Visual Evaluations results : Accuracy.}
    Comparison with ablations of our method discussed in Sec. \ref{supp_ablation} on all the distribution shifts used in the paper along with transfer learning experiments for the CLIP \textit{ViT-L/14@336px} model . The last column shows the average accuracy over all the shifts. The first row of numbers correspond to the version of our method without KL divergence term at inference time. Similarly, the row below it corresponds to our method but without uncertainty ($\mathcal{U}_{mc}$) term at inference time. The second last row corresponds to the variant of our method without any knowledge distillation and the last row correspond to our complete method using single modal ViTB/16 teacher.
    }
    \label{tab:ablation_table}
\end{table*}

     

\begin{table}[t]
    \centering
    \footnotesize
    \begin{tabular}{lc|ccccc}
    \toprule
       & & \multicolumn{5}{c}{{Distribution Shifts}}  \\
       & IN  & IN-R & IN-Sketch & ObjectNet & IN-A  & IN-C \\  
    \midrule
    \textit{w/o KLD} & 0.89 &  0.82 & 0.71 & 0.82& 0.79&  0.83\\
    \textit{w/o $\mathcal{U}_{mc}$} & 0.71 &  0.69 & 0.72 & 0.61 & 0.63& 0.71\\
    \midrule
    \textit{Ours}  & \textbf{0.95} & \textbf{0.94}& \textbf{0.89}& \textbf{0.84}& \textbf{{0.92}}& \textbf{0.94}\\
     \bottomrule
     
    \end{tabular}
    \vspace{0.1in}
    \caption{ \textbf{Visual Evaluations results : $F_{correct}$}.
    Analysis of our method and its ablations (discussed in Sec. \ref{supp_ablation}) using the $F_{correct}$ (fraction of examples for which correct head is selected) metric descirbed in Sec. \ref{supp_ablation}, for the CLIP \textit{ViT-L/14@336px} model. Here, \textit{w/o KLD} denotes the ablation without the KL divergence at inference time and similarly \textit{w/o $\mathcal{U}_{mc}$} denotes the ablation without the uncertainty term ($\mathcal{U}_{mc}$) during inference. For our method single modal ViT-B/16 is used as teacher.
    }
    \label{tab:ablation_table_fraction}
\end{table}

\begin{table}[]
    \centering
    \footnotesize
    \begin{tabular}{cc|c|cc}
        \toprule
        Student & Teacher &  \textit{Ours w/o K.D.} & Ours \\
        \midrule 
         RN-101 & RN-34 &  53.8 & 55.3 \\
         CLIP RN-101 & RN-101  & 54.9 & 56.7 \\ 
         CLIP ViT-B/16 & ViT-S  & 55.1 & 57.4 \\
         CLIP ViT-B/32 & ViT-S  & 54.5 & 57.1 \\
         \bottomrule
    \end{tabular}
    \vspace{0.1in}
    \caption{ \textbf{Knowledge Distillation Ablation : Accuracy}.
    Comparison with the \textit{w/o K.D.} (without knowledge distillation) ablation of our method (refer Sec. \ref{supp_ablation}) using both single and multi-modal (CLIP) networks as students and single modal teachers, under a Visual Evaluation setup on the ImageNet-C dataset.}
    \label{tab:ab_kd}
\end{table}

\section{Analyzing improvements in Teachers v/s Students}
\label{teach}
We now analyze the effect of improving robust accuracy of the small teacher model on the robustness it transfers to the student model using our method. For this, we analyze the ImageNet-C accuracy of the robustified student when using different teachers (having different ImageNet-C accuracies after their robustification). 
We fix the student to be a CLIP ViT-B/16 model and use robustified RN-34, RN-50, RN-101, RN-151, ViT-Tiny, ViT-Small as the teachers. Table \ref{tab:teacher_for_student} shows the results for this analysis where each row corresponds to a teacher and columns show the robust accuracy of both teacher and the robustified student on the ImagNet-C dataset. In majority cases difference in student accuracy between a given row and the row just upper it,
is higher as compared to the same quantity but for the teacher column. This shows that when we switch to a better teacher (\textit{i.e.} teacher accuracy gets increased) then the increment in student accuracy, in majority cases, is even more.

\begin{table}[]
    \centering
    \footnotesize
    \begin{tabular}{c|c|cc}
    \toprule
    \multicolumn{4}{c}{CLIP \textit{ViT-B/16} Student (149M)}\\
    \midrule
         &Params&  {Teacher} & {Student}
          \\
         \midrule
         RN-34 & 11M &  55.7 & 54.3\\
         RN-50 & 23M & 57.6 & 56.9\\
         RN-101& 45M &  58.7& 57.2\\
         ViT-Small& 48M& 60.9 & 57.4\\
         RN-151&60M & 62.4 & 59.7\\
         ViT-Base& 88M& 63.2 & 62.5\\
         \bottomrule
    \end{tabular}
    \vspace{0.1in}
    \caption{\textbf{Visual Evaluation Results : Accuracy}. Analyzing how robust accuracy of CLIP ViT-B/16 student changes with increasing robust accuracy of the teacher. Here, each row corresponds to a robustified teacher with architecture given in the first column.
    The teacher column shows the accuracy of this robustified teacher on the ImageNet-C data. Similarly, each student column element shows the accuracy of the fixed student on the ImageNet-C data, when the teacher corresponding to its row is used in our method to distill knowledge.}
    \label{tab:teacher_for_student}
\end{table}

\begin{table*}[!htb]
    \centering
    \footnotesize
    \begin{tabular}{c|cccccccccc}
    \toprule
    & RN-101 & RN-151 & ViT-S & ViT-B16 & RN-50C & RN-101C & ViT-B16C & ViT-L14C & LiT-B16/B\\
    \midrule
       Tuned & 2.1M & 2.6M & 2.3M & 3.1M & 3.9M & 4.1M & 4.3M & 5.6M & 4.4M  \\
       Total  & 45M & 60M &  48M & 88M & 102M & 119M &149M & 450M & 195M &  \\
         
    \end{tabular}
    \vspace{0.1in}
    \caption{Number of parameters tuned and total number of parameters for all the students used. Here, each column corresponds to a student network architecture used in this work. First row shows the number of parameters for a given column as student, when our method is applied to update this student. The last row shows the total parameter count of this student.}
    \label{tab:param_count}
\end{table*}

\section{Further results on Transfer Learning}
\label{transfer}
\begin{figure}
    \centering
    \includegraphics[width=0.9\linewidth]{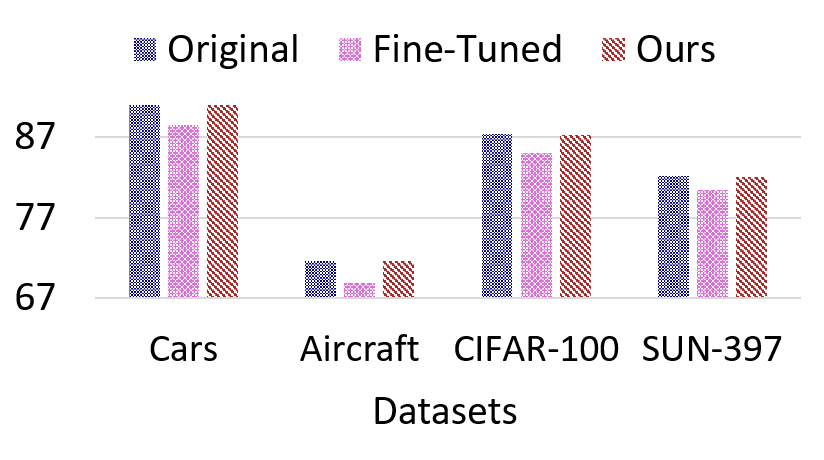}
    \caption{Transfer Learning under Visual Evaluation setup. The x-axis consist of various datasets and y-axis shows the accuracy on each of these datasets under the transfer learning setup. This analysis is done for original pre-trained CLIP \textit{ViT-L/14@333px} model (Original), its ImageNet fine-tuned version (Fine-tuned) and after it has been updated using our approach with a ViT-B/16 teacher (Ours).}
    \label{fig:transfer_supp}
\end{figure}

We further analyze our proposed scheme for the transfer learning setup under more datasets used in the CLIP paper for classification. Specifically, we use the Cars \cite{krause20133d}, CIFAR-100 \cite{krizhevsky2009learning}, Aircraft \cite{maji13fine-grained} and SUN-397 datasets \cite{xiao2010sun} for this setup and compare our model against the transfer learning of the original CLIP model and its completely fine-tuned version under the Visual Evaluation setting. Figure \ref{fig:transfer_supp} shows the results for this analysis where original denotes the visual encoder of the initial CLIP model without any tuning. It can be observed, similar to the main paper, that complete fine-tuning is not able to preserve the transfer learning or generalization capabilities of the model whereas our method preserves this important characterstic of a pretrained model.

\section{Analyzing more Multi-Modalities}
\label{mm}
\begin{figure}
    \centering
    \includegraphics[width=\linewidth]{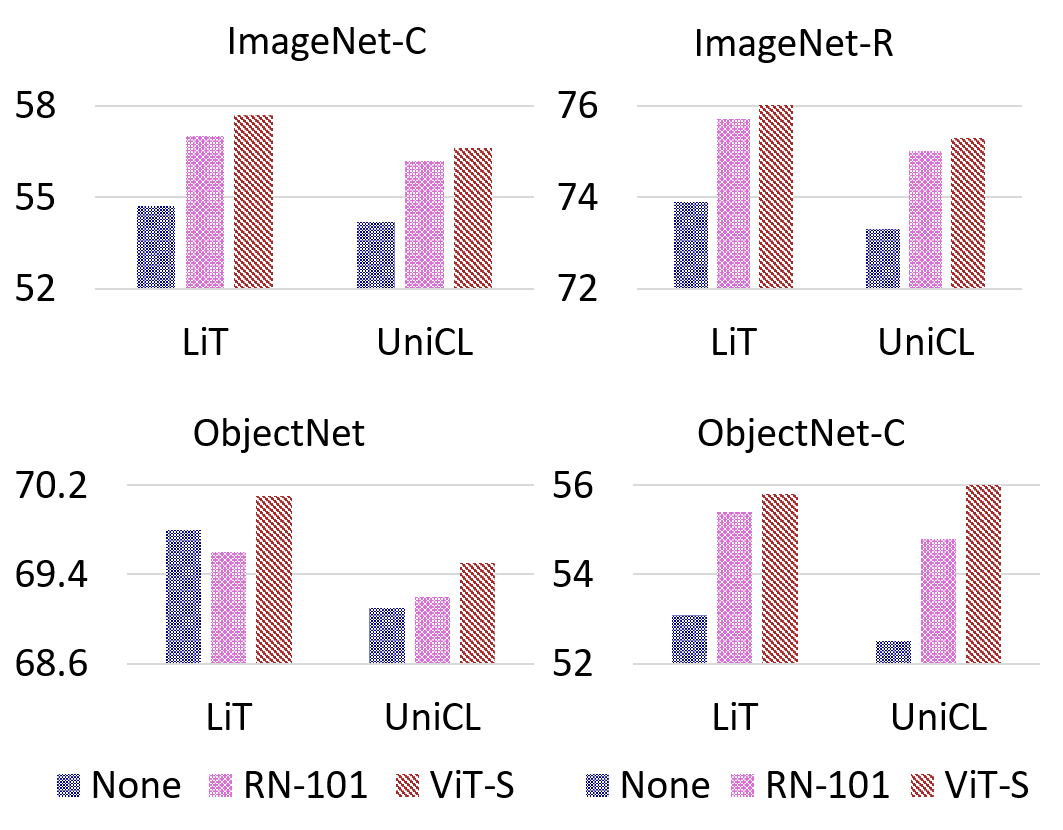}
    \caption{\textbf{Visual Evaluation Results : Multi-Modalities}. y-axis : Accuracy, x-axis : Multi-Modal Student, vertical bar : Teacher.This figure analyses our method on the ImageNet-C,R and ObjectNet,ObjectNet-C datasets when LiT-B/16B, UniCL SWIN-T multi-modal networks are used as students (x-axis) and are tuned using single modal networks (RN-101, ViT-S) as teacher (vertical bar). Vertical bar labelled None correspond to APT baseline. It can be observed that out method provides significant gains over the APT baseline, especially when ViT-S model is used as the teacher.}
    \label{fig:mm_lpl}
\end{figure}

We further analyze our method for other popular pretrained Multi-Modal Networks namely LiT\cite{zhai2022lit} and UniCL\cite{yang2022unified}. For LiT, we use the LiT-B/16B model and for UniCL we use the SWIN-T model.
Figure \ref{fig:mm_lpl} shows the results for various teacher student pairs under Visual Evaluation setting on ImageNet-C,R datastes and under the Zero Shot setting on ObjectNet,ObjectNet-C datasets. For both settings Unimodal ResNet-101 and ViT-Small are used as teachers. 
Each vertical bar for a given student on x-axis (LiT/UniCL) correspond to a particular teacher and y-axis denotes the accuracy. 
For Visual Evaluation, we use APT baseline (rows with Teacher set to None) and for Zero-Shot, original Zero-Shot model (without any fine-tuning) is used as the baseline.\\
It can be observed that our method again improves the accuracy over the APT baseline by around 2.5\% (average) on the ImageNet-C dataset and by around 1.9\% (average) on the ImageNet-R dataset under the Visual Evaluation setting. Similarly for Zero Shot setting, it improves accuracy by around 2.5\% (average) over the baseline on the ObjectNet-C dataset. This shows that our method generalizes well to other pretrained multi-modalities as well.

\section{Analysis on ImageNet-P}
\label{im_p}
\begin{figure}
    \centering
    \includegraphics[width=0.9\linewidth]{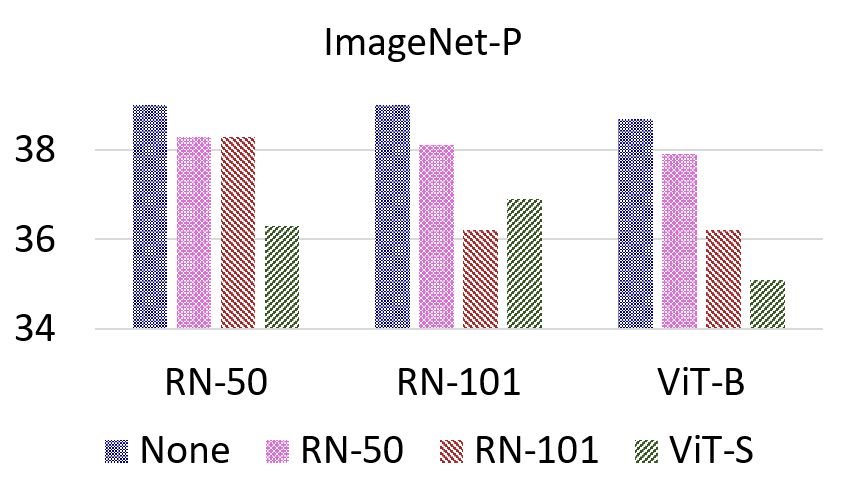}
    \caption{\textbf{Visual Evaluation Results : mFR}. y-axis : mFR, x-axis : CLIP Student Network architectures, vertical bar : Teacher. This figure analyses our method on the ImageNet-P dataset when various CLIP model architectures are used as students (x-axis) and tuned using various single modal networks as teacher (vertical bar). Vertical bar labelled None correspond to APT baseline. It can be observed that out method provides significant gains over the APT baseline, especially when ViT-S model is used as the teacher.}
    \label{fig:im_p}
\end{figure}

We further analyze robustness using the mean Flipping Rate (mFR)\cite{hendrycks2019benchmarking} for the ImageNet-P dataset for various CLIP models (RN-50,101 and ViT-B/16) under the Visual Evaluation setting. \\
We begin by first analyzing the robustness of the teacher models, robustified using the scheme described in the main paper (using AugMix+DeepAugment). Table \ref{tab:augmix} shows this analysis, comparing the mFR metric for these single-modal teacher networks, with and without (Naive) applying the robustification scheme. It can be observed that their performance is significantly improved after robustification. The best performance is shown by the ViT-Small model. Even though the model has comparable parameters to ResNet-101, still a significant difference in the performance highlights its robust learning scheme. Given the robustified teachers, we now evaluate students tuned using our method on this dataset.
Figure \ref{fig:im_p} shows the results for the CLIP model students using our method and also for the APT baseline. It can be observed that our method is able to reduce mFR by upto 3\% and greater than 2\% for the cases with ViT-Small as the teacher. Even thouh ViT-Small and RN-101 have similar number of parameters, still a significant difference is observed for CLIP RN-50 and CLIP ViT-B/16 model showing that ViT transfers the robustness more efficiently w.r.t. this dataset.


\begin{table}[]
    \centering
    \footnotesize
    \begin{tabular}{c|cc}
        \toprule
        Arch.  &  Naive &   AugMix+DA \\
        \midrule
        RN-34 & 59.8 & 41.6 \\ 
        RN-50 & 57.1 & 37.5\\ 
        \midrule
        RN-101 & 52.7 & 34.6\\ 
        ViT-Small & 49.2 & 33.4\\ 
         \bottomrule
    \end{tabular}
    \vspace{0.1in}
    \caption{Comparison of mean Flipping Rate (mFR) on ImageNet-P dataset to analyze  robustness induced by applying AugMix+DA for tuning the relatively small teacher models.}
    \label{tab:augmix}
\end{table}


\section{Parameters Tuned}
\label{param}
We provide the number of parameters tuned for each network used as student and updated using our approach along with their total parameter count in table \ref{tab:param_count}. Unless mentioned the number of parameters tuned for a given student in all of the experiments corresponds to the value provided in this table.

\section{Implementation Details}
\label{impl}
The network tuning using our algorithm involves updating a portion (refer table \ref{tab:param_count}) of the complete architecture starting from end using a learning rate of $1e-3$ and a batch size of 256 using an Adam optimizer. For tuning our method, unless mentioned, we use the half of the complete augmented data (a=0.5), generated by DeepAugment+Augmix. The tuning is carried out for 500 epochs. Robustifying the teacher involves the same pipeline and hyper-parameters as proposed in the DeepAugment paper\cite{hendrycks2021many}.  Also a dropout is applied while training with probability set to 0.25. The last one-fifth of the tuning portion corresponds to clean/unclean and combined heads. At the inference time, $N=10$ samples are drawn per head with dropout activated to estimate uncertainty term ($\mathcal{U}_{mc}$). Not, this requires multiple passes only through dropout activated layers, not the complete model.

\end{document}